\documentclass{article}
\usepackage{ijcai18}

\usepackage{times}
\usepackage{xcolor}
\usepackage{soul}
\usepackage[utf8]{inputenc}
\usepackage[small]{caption}

\usepackage{graphicx}
\usepackage{subfig}

\usepackage{amsfonts}
\usepackage{amsmath}
\usepackage{breakcites}
\usepackage{amssymb}
\usepackage{bm}

\usepackage{algorithm}
\usepackage{algorithmic}
\usepackage{setspace}

\usepackage{multirow}

\usepackage{enumitem}
\setlist{nolistsep}

\usepackage{graphicx}
\usepackage{subfig} 

\usepackage{color}
\newcommand{\wtw}[1]{\textit{\color{red}#1}}

\title{Disentangling Aspect and Opinion Words in Target-based Sentiment Analysis\\ using Lifelong Learning}

\author{
Shuai Wang$^1$, 
Mianwei Zhou$^2$, 
Sahisnu Mazumder$^1$, 
Bing Liu$^1$,
Yi Chang$^3$,
\\ 
$^1$ University of Illinois at Chicago \\
$^2$ Yahoo! Research \\
$^3$ Huawei Research America  \\
shuaiwanghk@gmail.com,
mianwei@yahoo-inc.com,
sahisnumazumder@gmail.com,\\
liub@cs.uic.edu,
yichang@acm.org
}

\begin{document}

\maketitle

\begin{abstract}
Given a target name, which can be a product aspect or entity, identifying its aspect words and opinion words in a given corpus is a fine-grained task in target-based sentiment analysis (TSA). This task is challenging, especially when we have no labeled data and we want to perform it for any given domain. To address it, we propose a general two-stage approach. Stage one extracts/groups the target-related words (call t-words) for a given target. This is relatively easy as we can apply an existing semantics-based learning technique. Stage two separates the aspect and opinion words from the grouped t-words, which is challenging because we often do not have enough word-level aspect and opinion labels. In this work, we formulate this problem in a PU learning setting and incorporate the idea of lifelong learning to solve it. Experimental results show the effectiveness of our approach.

\end{abstract}

\section{Introduction}
Target-based sentiment analysis (TSA) is an important topic in sentiment analysis~\cite{jiang2011target,vo2015target,wang2016targeted}. A target can be a product aspect or entity, and is also referred to as \textit{aspect} in the literature. In this paper, we use \textit{target} as an aspect category and \textit{aspect words} as related mentions of the target. We focus on a fine-grained TSA (or FTSA) task: \textit{given a target name, identifying its aspect words and opinion words in a given domain corpus}. For example, one is interested in opinions about the target ``screen'' of a camera, and wants to find out all related aspect and opinion words mentioned in reviews. We may find aspect words like ``LCD,'' ``display'' and ``resolution,'' and opinion related words like ``scratched,'' ``blurry'' and ``bubbly.''

This problem is challenging, especially when there is no labeled data and we want to perform it in any domain (reviews of any product). In practice, it is not feasible to manually annotate all possible targets beforehand for every domain. Designing an unsupervised or semi-supervised method for the task is thus needed. For this goal, we developed a two-stage approach which does not require manual labeling.


\begin{figure}[t]
\label{fig:first}
\centering
\subfloat[Grouping]{
\includegraphics[scale=.32]{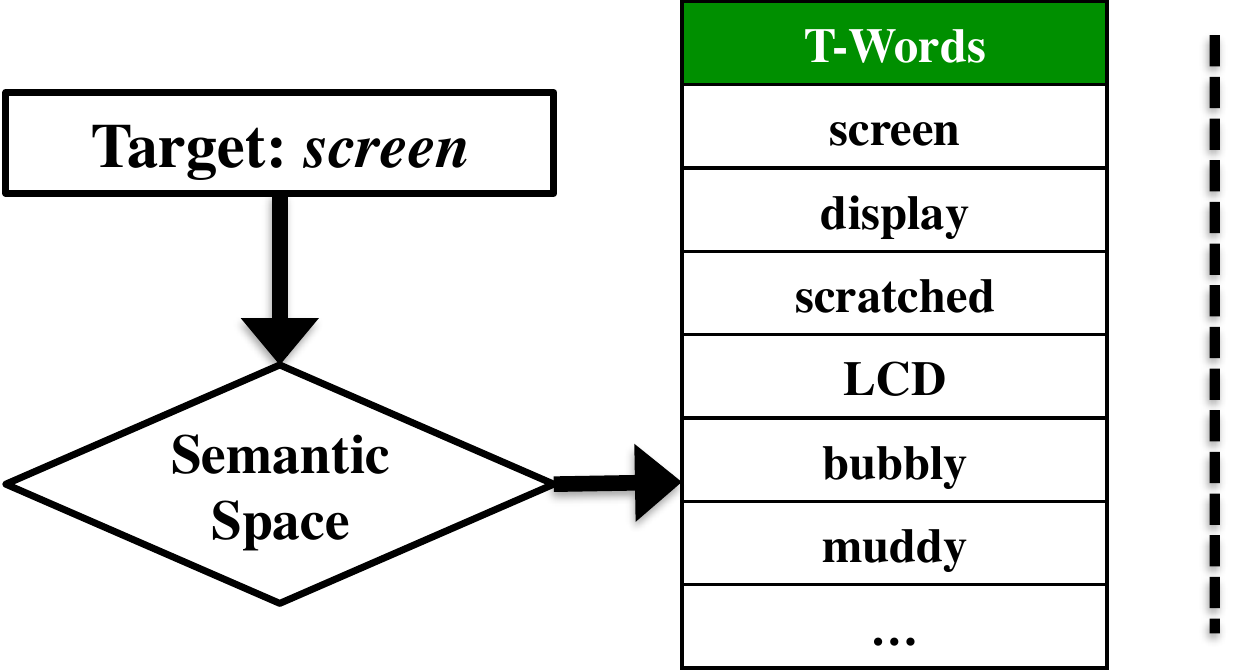}
\label{fig:keyword}
}
\subfloat[Disentangling]{
\hspace{0.2mm}
\includegraphics[scale=.32]{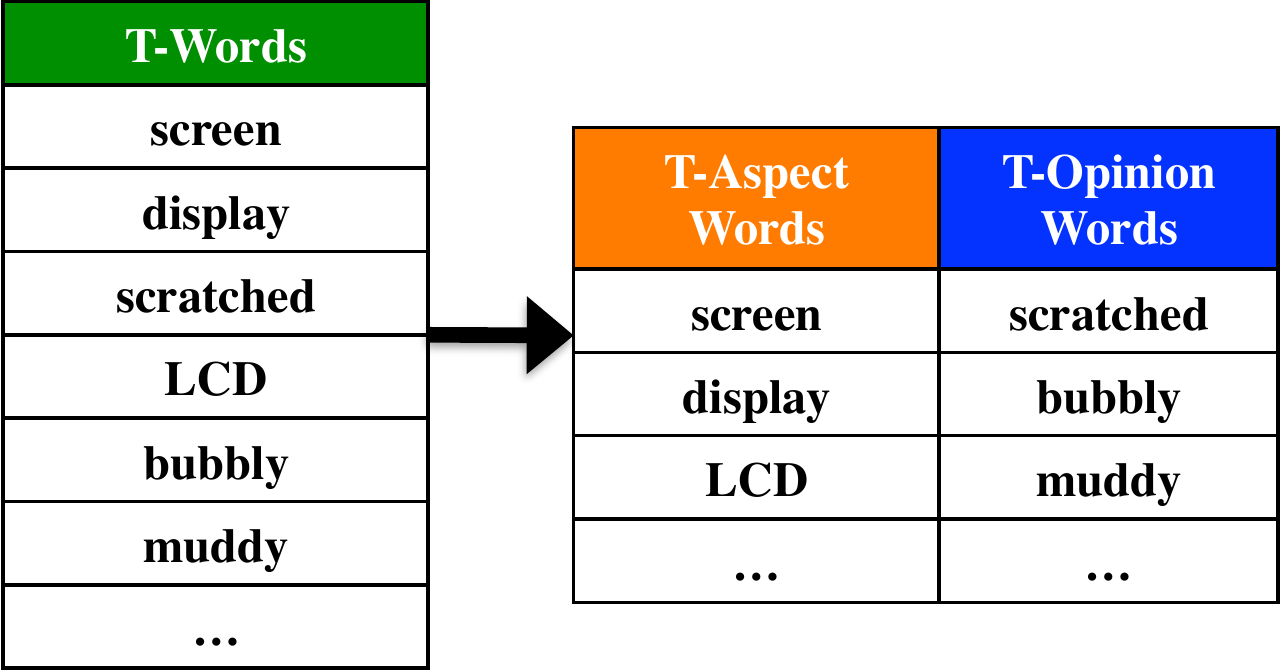}
\vspace{-10mm}
\label{fig:example}
}
\caption{Two-stage approach to Fine-grained TSA (FTSA)}
\vspace{-5mm}
\end{figure}

Stage one is defined as target-related words extraction/grouping. That is, given a target name, we first identify the target-related words (called \textit{t-words}) in a corpus. For instance, when the target is \textit{screen}, the t-words ``display'', ``LCD'', ``scratched'' and ``bubbly'' are extracted (as shown in  Figure~\ref{fig:keyword}). We can achieve this by using the semantic representation of words obtained from distributional representation learning~\cite{levy2015improving}. Specifically, given a target name, we extract its semantically similar words from its semantic representation as t-words. This grouping stage is very similar to the (unsupervised) aspect extraction in aspect-level sentiment analysis, so many existing approaches~\cite{mukherjee2012aspect,liu2012sentiment} exploiting the learned semantics can be utilized too, like topic modeling~\cite{blei2003latent}. 


One key issue of most semantics learning techniques for our task (FTSA) is that, they will inevitably couple the target-related aspect words (called \textit{t-aspect words}) and opinion words (called \textit{t-opinion words}). We interpret its cause from a linguistic perspective. Most semantics learning models are developed based on the idea of \textit{distributional hypothesis}: linguistic items occurring with similar contexts have similar meanings~\cite{harris1954distributional}, so they in fact group two different types of semantic similarity together, namely, conceptual and associative similarity. Conceptual similarity means two words are conceptually similar (likely replaceable), like ``dog'' and ``canine''. Associative similarity means two words tend to appear in similar contexts, like ``dog'' and ``bark.'' The distinction between them is well-known in cognitive science~\cite{tversky1977features}, and it has also been discussed in NLP~\cite{kiela2015specializing,levy2015improving}. In regard to sentiment analysis, we can see that t-aspect words ``display'' and ``LCD'' and t-opinion words ``scratched'' and ``bubbly'' are all mixed based on our given example (Figure~\ref{fig:keyword}).

In spite of the discussed drawback, we argue that the semantics-based models are still suitable to achieve our final goal (i.e., FTSA), with the reason being three-fold. First, the mixture benefits the t-words grouping, where both two types of semantic correlation can be jointly extracted. To be concrete, aspect words like ``display'' could be found because of the conceptual similarity (similar to ``screen'') and opinion words like ``scratched'' could also be discovered due to associative similarity (associated with ``screen''). Second, those existing or emerging aspect extraction models can be utilized, which paves the way for better FTSA. Third, those semantics-based models are usually learned in an unsupervised or semi-supervised manner, which meets our need. However, when we take advantage of those semantics-based models for stage one, we have to overcome their aforementioned drawbacks for performing FTSA, which leads to our stage two.


Stage two is defined as: \textit{Given a list of target-related words (\textit{t-words}), separating them into target-related aspect words (\textit{t-target words}) and target-specific opinion words (\textit{t-opinion words})}. Figure~\ref{fig:example} shows an example. Notice that the list of t-words is assumed to be given, which is grouped by an existing semantics-based learning technique, so we refer to this problem as \textbf{\textit{disentangling aspect and opinion words from extracted/grouped target-related words}}, to distinguish it from other related sentiment analysis problems (see Section 2). 


An intuitive solution to this problem is to model it as a word-level binary classification task. That is, to build a classifier for learning and predicting t-aspect and t-opinion words. However, this is difficult in practice, because this means that we need both aspect and opinion word-level labels for every domain, which requires intensive human efforts. Noticing this, we instead formulate the classification problem in a PU (Positive-Unlabeled) learning setting. The idea is to use general/common opinion words (treating them as positive examples) to distill other opinion words from unlabeled words. However, a notable issue in this PU setting is that the errors from false positive (FP) examples (wrongly predicted opinion words) can be propagated, resulting in more errors and degenerating its performance. To address this issue, we exploit the idea of \textit{lifelong machine learning}~\cite{thrun1998lifelong, chen2016lifelong} and incorporate it into the PU learning process. We name it as Lifelong PU learning (LPU). It works by accumulating the knowledge learned from (past) multiple domains, and uses it to restrict the propagation of FP examples and to ensure the reliability of the newly learned opinion words. Our experimental results show its effectiveness.

The main contributions of this paper are summarized as follows: (1) It proposes to perform the fine-grained target-based sentiment analysis (TFSA) task in a two-stage manner, which does not require manual labeling. (2) It proposes a lifelong PU (LPU) learning approach to solving the problem of disentangling target-specific aspect and opinion words from word extraction/grouping.  To the best of our knowledge, none of the existing studies has employed the lifelong PU technique. (3) Experimental results conducted on real-world review datasets with two general aspect extraction techniques show its effectiveness and extensibility.

\section{Related Work}
\label{sec:relatedwork}
\noindent
\textbf{Target-based sentiment analysis}
Target-based sentiment analysis (TSA) aims at analyzing the sentiment on a specific given target. Most of the previous studies~\cite{jiang2011target,vo2015target} focused on the target-dependent sentiment classification task, which is to classify the sentiment polarity on a sentence towards a given target, for example, to determine whether a tweet from Twitter shows positive or negative sentiment towards a company. Our task is not to classify a single sentence, but to identify all aspect and opinion words in a given corpus with a specified target name. Wang et al.~\shortcite{wang2016targeted} proposed a targeted topic model to generate target-related topics. However, their work dealt with neither opinions nor the word disentangling problem. 

\noindent
\textbf{Aspect-level sentiment analysis}
Our work is also related to the widely studied aspect-level sentiment analysis (ASA) ~\cite{liu2012sentiment}. To show the difference and relatedness of ASA to our problem (FTSA), we categorize previous studies into three general groups. The first group uses linguistic tools, association structures or hand-crafted rules for aspect or opinion word extraction, or co-extraction~\cite{hu2004mining,qiu2009expanding,xu2013mining}. The second group models the aspect and opinion word identification as a supervised sequence labeling problem~\cite{irsoy2014opinion,wang2017coupled}. These two groups require intensive human labors for feature engineering, pattern design or manual labeling. The third group, in contrast, does not rely on human involvement, and it exploits the distributional semantics, e.g., topic modeling, word embedding, or their variants ~\cite{mukherjee2012aspect,wang2016mining,tixier2016word}. We have discussed the suitability of these techniques for our task in Section 1. Note that our focus is on disentangling (separating) t-aspect and t-opinion words from the t-words grouped by them. Some of the studies in the third group also considered separating the aspect and opinion words, by using POS features with manual or auto labels. It is a purely syntax-based solution. We will compare and analyze it in our experiments. Most importantly, most existing methods performed full analyses on all aspects, while our task is target-oriented.

\noindent
\textbf{Semantic space and representation} Semantics-based learning models project words to a semantic space and represent each word as a dense vector. Such semantics-bearing vectors can be created by matrix factorization (e.g. LSI)~\cite{deerwester1990indexing} and topic modeling (e.g., LDA)~\cite{blei2003latent}. Their word vectors have been used to tackle some word-level classification problems~\cite{maas2011learning,pu2015topic}. Recently, neural word embeddings~\cite{mikolov2013distributed,pennington2014glove} emerge to show better semantic representation for words and have improved many NLP tasks~\cite{turian2010word,collobert2011natural}. 


\noindent
\textbf{Lifelong machine learning}
Our work is also related to lifelong learning~\cite{thrun1998lifelong,chen2016lifelong}. In the sentiment analysis context, several lifelong learning models have been proposed for improving topic quality~\cite{wang2016mining}, aspect extraction~\cite{liu2016improving} and document-level sentiment classification~\cite{chen2015lifelong}. But they are not for the TSA task and they are not applicable to the word disentangling problem. Additionally, we need to formulate our problem in a PU learning setting and we use no labeled data. We also incorporate lifelong learning into the PU learning process. To the best of our knowledge, none of the previous studies have employed the lifelong PU technique.

\section{Stage one: Grouping} 
\label{sec:grouping}
In this stage, we group the target-related words (t-words) for a specified target name. The basic idea is to extract its semantically correlated words based on the vector representation of the target in a learned semantic space. Specifically, we use the neural word embedding model~\cite{mikolov2013distributed} to learn word vectors for a given domain corpus, resulting in an embedding matrix $E \in \mathbb{R}^{v \times d}$ where $v$ and $d$ are the size of vocabulary and vector dimension. Then a semantic similarity matrix $M \in \mathbb{R}^{v \times v}$ is calculated based on the dot product of $E$ and $E^T$. After that, when a user-specified target comes, the nearest neighbors of the target word will be returned as t-words, based on their similarity values 
in $M$. Notice that other semantics learning models can be used in the same way~\cite{deerwester1990indexing,pennington2014glove}. Probabilistic topic models~\cite{blei2003latent} can be used as well, by searching the corresponding topic for the given target and returning the topical words. Notice that this stage is to some extent similar to (unsupervised) aspect extraction in aspect-level sentiment analysis~\cite{liu2012sentiment} and many of its models can be used at this stage. That main difference is that, in our setting, the target/aspect name is specified, so we do not have to perform a full extraction of all aspects covered by a given corpus (e.g., performing clustering in word embedding space), but only to focus on the given target by returning its nearest neighbors.

\section{Stage Two: Disentangling}
\subsection{PU Learning using Word Vectors} 
This stage separates the given t-words into \textit{t-aspect words} and \textit{t-opinion words}. As discussed in Section 1, in order to provide a general approach without manual labeling for every domain, we formulate this disentangling problem as a binary classification task in a PU learning setting~\cite{li2005learning}.

Clearly, in addition to aspect and opinion words, a domain vocabulary also contains other words like general/background words. However, as indicated in~\cite{mukherjee2012aspect,wang2016mining}, those words do not have a seriously bad effect as they are unlikely to be semantically similar to a given target. Therefore, we assume most of the non-opinion words that are semantically correlated to a given target, are aspect words. This assumption holds well in practice, as shown in previous studies~\cite{mukherjee2012aspect,wang2016mining} and also in our experiments. 

PU learning is a type of semi-supervised learning method, which learns a binary classifier using only positive and unlabeled examples (with no negative examples). Here $P$ represents a set of data examples with positive labels. In our task, the opinion words from an opinion lexicon will be the words in $P$, such as ``good", ``bad'' and ``angry''. In terms of $U$, it denotes the set of data examples with unknown labels. In our case, other words that are not in the lexicon are in $U$. Note $U$ in fact contains both true opinion words and non-opinion words. With word vectors as features and a set of general opinion words as positive labels, we can build a PU classifier. In our work, we use logistic regression for classification, as it generates a probabilistic score of a word for being in the positive class (i.e., opinion word). In this way, some words from $U$ with high prediction scores can be found as new (likely) opinion words, and we can extract more words iteratively using the PU classifier.

However, a notable issue in this PU setting is that the errors from false positive (FP) examples (wrongly predicted opinion words) can be propagated, thus degenerating its performance. In order to address it, we exploit the idea of \textit{lifelong machine learning}~\cite{thrun1998lifelong, chen2016lifelong} and incorporate it into the PU learning process. The idea is to exploit the past domain classification knowledge to increase the correctness or reliability of the newly found opinion words. 

\subsection{Lifelong Machine Learning} 
Lifelong machine learning~\cite{thrun1998lifelong, chen2016lifelong} or \textit{lifelong learning} for short, works by retaining the knowledge learned from the past tasks and uses it to help future learning, i.e., to help the current or coming task. It mimics how we human beings learn. With regard to sentiment analysis, we (human beings) can learn many opinion expressions in our lives across different domains/areas, which enables us to better understand and identify opinion words in a new domain.

In a similar way, our system retains the newly learned opinion words every time it has finished processing one domain (one task), treating them as knowledge and accumulating them in a knowledge base. The system accumulates such knowledge continuously from domain to domain. So in any time it has processed $N$ domains and starts to process the $(N$$+$$1)$th domain, the accumulated knowledge will be used to help generate more reliable opinion words that are suitable for the $(N$$+$$1)$th domain. Based on this general idea, we develop a lifelong PU (LPU) learning algorithm.

\subsection{Lifelong PU learning (LPU)}
Our proposed LPU algorithm consists of four main steps: knowledge accumulation, current domain setup, knowledge mining and preparation, and restricted PU iterations. The overall algorithm is given in Algorithm~\ref{alg:LPU_learing} (Alg.~1). 

\noindent
\textbf{Step 1: Knowledge Accumulation (lines 1-8)} This step follows the traditional classification process but with knowledge retention for building a knowledge base from past domains. Specifically, for each domain (task $j$), we first obtain its vocabulary $W_j$ and semantic representation of words $V_j$ (line 3). With a general opinion lexicon, we then have the lexicon-based opinion words $W_{j}^P$, i.e., positive examples, and unlabeled examples $W_{j}^U$ (line 4). A PU classifier is trained (line 5) and used to predict the probabilistic class scores of words in $W_{j}^U$ and to find new opinion words $W_{j}^+$ (line 6). After that, we retain $W_{j}^+$ as knowledge for constructing a knowledge base $SKB$ (line 7).

Notice that in practice, we do not need to repeat this step every time we have/start a new domain/task. Instead, this step is performed naturally and continuously with domains being processed, from task 1 to task $N$. Because we simply keep retaining their results, so when a new domain comes (task $N+1$), the $SKB$ is already constructed and ready for use. 



\textbf{Step 2: Current Domain Setup (lines 9-13)} This step is for the setup of processing the current domain. The vocabulary words $W_i$ and their semantic representation $V_i$, lexicon-based opinion words (positive examples) $W_i^{p}$ and unlabeled words $W_i^{U}$ of the current domain are first obtained (lines 10-11). Then we build a hash table $H$ to store the nearest neighbors\footnote{Simply using top 10 neighbors works consistently well for different domains in our experiments.} for all words, which can be easily constructed from the similarity matrix $M$ (see Section \ref{sec:grouping}). With the table $H$ established (line 12), which is a one-time effort, the similarity query becomes a lookup operation. This $H$ not only helps in the current step 2, but also plays a role in the following step 4, as we will see shortly. Based on $H$, we can find the nearest neighbors for the lexicon-based opinion words and we call it reliable neighbors (line 13). This is an initial constraint, which is also intuitive, as the candidate/unlabeled words similar to the opinion words from the lexicon are believed to be more reliable/likely opinion words.

\textbf{Step 3: Knowledge Mining and Preparation (lines 14-19)}. This step is for mining knowledge and making preparation for later use. With the knowledge accumulated from many past domains and stored in $SKB$, we can extract the reliable knowledge $W^{SK}$ (line 15). Here we adopt the data mining technique of frequent itemset mining (FIM)~\cite{agrawal1994fast}, because a candidate word that frequently appears in many different domains as a predicted opinion word is naturally more trustworthy. The intersection of the reliable neighbors $W_{i}^{RN}$ and reliable knowledge $W_{i}^{SK}$ initializes $W_{i}^{NS}$, the newly learned sentiment (line 19). 
Lines 16-18 define other variables that are used in step 4, where $W_{i}^{SK}$ denotes the sentiment knowledge for current domain $i$, $W_{i}^{RS}$ indicates the reliable learned sentiment (opinion words) during the PU learning iteration, and the $W_{i}^{PP}$ records the newly-predicted opinion words in an ongoing iteration. 

\textbf{Step 4: Restricted PU Iterations (lines 21-31)}. This step performs iterations of PU learning with constraints. Unlike the unconstrained self-bootstrapping approach, the expansion of the newly-predicted opinion words as positive examples in LPU is controlled and only the reliable ones will be used further. The initialized new opinion words $W_{i}^{NS}$ have already been restricted (see step 2) and used here as initial reliable sentiment $W_i^{RS}$. During the iterative learning, it keeps being updated (line 23) by adding only reliable opinion words (line 27). We develop two ways of expanding new reliable opinion words. One way is to learn from the reliable knowledge (line 25) and another way is to learn from its self-predicted results (line 26). Notice that both ways of expansion are restricted by the defined reliability score shown in Algorithm~\ref{alg:reliableSentimetnt} (Alg.~2). This score is calculated based on the number of identified positive neighbors of a candidate word, which is also used for ranking. In Alg.~\ref{alg:reliableSentimetnt}, $A$ denotes the candidate word set and $B$ records positive examples. The identified positive neighbors are from the intersection of positive examples (provided by $B$) and the neighbors of a candidate word (provide by $H(w)$). In each iteration, only the top $l$ ranked words will be trusted/added as new positive examples. When the maximum iteration is met or there is no more new opinion words that the system can learn, the iterative learning process stops and all newly-detected opinion words are returned (line 31).

\begingroup
\begin{algorithm}[t]
   \caption{\small{Lifelong PU (LPU) Learning } }
      \label{alg:LPU_learing}
   \begin{tabular}{rl}
\textbf{Input}:& Current domain corpus $D_{i=n+1}$ \\
               & Past domain corpora $\bm D$=\{$D_1, .. D_{j}.., D_n$\}\\
               & Opinion words in lexicon $W^P$\\
               & Maximum learning iteration $m$  \\
               & Number of learned words in one iteration $l$\\   
\textbf{Output}:& All newly-extracted opinion words $W_{i}^+$ in $D_{i}$
\end{tabular}
\begin{spacing}{0.85}
\begin{algorithmic}[1]
\renewcommand{\algorithmiccomment}[1]{#1} 
\STATE \COMMENT{// Step 1. Knowledge Accumulation}
	  \FOR{each domain corpus $D_{j} \in \bm D$}
	\STATE $W_j, V_j \leftarrow GetWordsAndEmbeddings(D_{j})$
	\STATE $W_j^P \leftarrow W_j \cap W^P$, $W_j^U \leftarrow W_j - W_j^P$
	\STATE $c_{j} \leftarrow PUClassifier(V_j, W_j^P)$
	\STATE $W_{j}^+ \leftarrow OpinionWordPrediction(c_{j}, W_j^U)$
	\STATE $\bm{SKB} \leftarrow \bm{SKB} \cup W_j^+$ // sentiment knowledge base
	\ENDFOR
\STATE \COMMENT{// Step 2. Current Domain Setup}	
	\STATE $ W_i, V_i \leftarrow GetWordsAndEmbeddings(D_i)$
	\STATE $W_i^P \leftarrow W_i \cap W^P$, $W_i^U \leftarrow W_i - W_i^P$
	\STATE Create a hash-table $H$ to store top neighbors of all words
    \STATE $W_i^{RN} \leftarrow ReliableNeighbors(H, W_i^P)$
\STATE \COMMENT{// Step 3. Knowledge Mining and Preparation}
	\STATE $ W^{SK} \leftarrow FIM(\bm{SKB})$
	\STATE $W_i^{SK} \leftarrow W_i \cap W^{SK}$ // senti-knowledge for domain $i$
	\STATE $W_i^{RS} = \varnothing$  // reliable learned opinion words
	\STATE $W_i^{PP} = \varnothing$  // current positive prediction (opinion words)  
	\STATE $W_i^{NS} = W_i^{RN} \cap W_i^{SK}$  // newly learned sentiment 
\STATE \COMMENT{// Step 4. Restricted PU Iterations}
	\STATE $t = 0$
	\WHILE{$t < m$ or $W_i^{NS}$ is not empty}
	\STATE $W_i^{RS}  \leftarrow  W_i^{RS} \cup W_i^{NS}$ // updating reliable sentiment
	\STATE $c_{i} \leftarrow PUClassifier(V_i, W_i^{RS} \cup W_i^{P})$
	\STATE $ W_i^{NEW1} \leftarrow ReliableOpinion({W_i^{SK}, W_i^{PP}, H}, l)$  
	\STATE $ W_i^{NEW2} \leftarrow ReliableOpinion({W_i^{PP}, W_i^{PP}, H, l})$  
	\STATE $ W_i^{NS} \leftarrow W_i^{NEW1} \cup W_i^{NEW2} $
	\STATE $W_i^{PP} \leftarrow OpinionWordPrediction(c_{i}, V_i)$
	\STATE $t = t+1$
	\ENDWHILE
		\STATE $W_{i}^+ \leftarrow OpinionWordPrediction(c_{i}, W_i^U)$
\end{algorithmic}
\end{spacing}
\end{algorithm}
\endgroup

\begingroup
\begin{algorithm}[tb]
   \caption{$ReliableOpinion({A, B, H, l})$  }
      \label{alg:reliableSentimetnt}
\begin{spacing}{1}
\begin{algorithmic}[1]
\renewcommand{\algorithmiccomment}[1]{#1} 
	\STATE $S = \varnothing$ // counts positive neighbors for every word in A
	  \FOR{each a word  $w \in A$}
	\STATE $S  \leftarrow {countPositiveNeighbors}(B, H(w))$
	\ENDFOR
	\STATE return ${sortAndReturnTopLWords}(A, S, l)$
\end{algorithmic}
\end{spacing}
\end{algorithm}
\endgroup

\section{Experiments}
\label{sec:experiments}
\subsection{Candidate Methods for Comparison}
\textbf{Adjective Extraction (ADJ): } This baseline simply regards all adjective words as opinion words. This is a simple but widely used solution. We performed POS tagging and extracted all adjectives. No classifier is used for ADJ. \\
\textbf{Part-Of-Speech (POS): } 
The POS features have been shown very effective for aspect and opinion extraction tasks. This is a representative syntax-based approach used in many related works~\cite{mukherjee2012aspect,wang2016mining}. Here every word is represented by the POS features of its context, i.e., $w_i$ will be represented as [$POS_{i-1}$,$POS_{i}$,$POS_{i+1}$]. It is used as word representation for building a classifier. \\
\textbf{Latent Semantic Indexing (LSI): } LSI is a standard matrix factorization technique to construct latent semantic vectors/features. Its factorized word-feature correlation matrix can be used as word vector representation~\cite{pu2015topic}. \\
\textbf{Latent Dirichlet Allocation (LDA): } LDA~\cite{blei2003latent} is a classic topic model which discovers hidden topics from documents and groups words into topics. Similar to LSI, the term-topic matrix is used as the word vector representation to build a classifier~\cite{maas2011learning}. \\
\textbf{Non-Lifelong Learning (NLL): } This baseline follows our approach but with no lifelong learning. It uses the word vectors learned by neural word embeddings to build a classifier.\\
\textbf{Lifelong PU (LPU): } This is our proposed lifelong PU learning algorithm introduced in Alg.~1. \\
\textbf{Lifelong PU minor (LPU-): } This is a LPU variant that does not make risky self-prediction exploration and relies more on the past mined knowledge. In other words, it considers the first type of reliable sentiment only but without the second one (see lines 25 and 26 in Alg.~1). This can be viewed as a conservative version of LPU. \\
\vspace{-5mm}


\subsection{Experimental Setup} 
\textbf{Data} We use a large corpus of Amazon reviews from 20 different domains provided by~\cite{mcauley2015inferring}{\footnote{http://jmcauley.ucsd.edu/data/amazon/}} and the full list is shown in Table~\ref{tab:dataset}. For training (all PU classifiers), a general opinion lexicon\footnote{http://www.cs.uic.edu/liub/FBS/sentiment-analysis.html} is used so the words appeared in it are automatically labeled as P. For testing/evaluation, we manually label the aspect and opinion words. Specifically, three domains from different product categories are selected, namely, \textit{cellphone}, \textit{beauty} and \textit{office}. For each domain, three targets are specified (see Table~\ref{tab:dataset}). In Table~\ref{tab:dataset}, the vocabulary size is the number of words after filtering the words with low occurrence (less than 5). We did not do any further preprocessing such as stemming or lemmatization, as the opinion words are also related to their grammatical forms.


\begin{table*}
\small
\begin{center}
\begin{tabular}{|c|c|c|c|c|c}
\hline
\rule{0pt}{8pt} \textbf{Dataset Name} & \textbf{Number of Reviews}   & \textbf{Vocabulary Size} & \textbf{Words in Lexicon}
& \textbf{Targets for Evaluation}\\
\hline
CellPhone & 194,439  &  28,942 & 2,764 & display, volumes, weight\\
\hline
Beauty  & 198,502  &  29,695 & 2,778 & cleansers, fragrance, groomers\\
\hline
Office  & 53, 258 &  20,858 & 2,332 & papers, clips, chairs\\
\hline
Full domain list & \multicolumn{4}{|c|}{\rule{0pt}{8pt}{apps for android, amazon instance video, automotive, baby, beauty,  cd, cellphone, cloth, digital music, electronics, }}\\
& \multicolumn{4}{|c|}{\rule{0pt}{8pt}{grocery, health,  kindle, tools/home improvement, home and kitchen, office product, pet supplies, sport, toy, video game}}\\
\hline

\end{tabular}
\end{center}
\vspace{-2.5mm}
\caption{\label{tab:dataset} Detailed information about the three domains for evaluation and the full domain list.}
\vspace{-6.5mm} 
\end{table*} 

\noindent
\textbf{Parameters and Settings}
For every candidate method except ADJ, their word vectors/features are learned and used for classification. Specifically, for LSI and LDA, we obtained the term-feature matrix and term-topic matrix using gensim package\footnote{https://radimrehurek.com/gensim/}. For NLL, LPU- and LPU, we used the skip-gram model~\cite{mikolov2013distributed}. The vector dimension is set to 200 as default and we maintain the same size for LDA and LSI. Logistic regression is used as the classifier in all methods. For LPU, we treat other 19 domains besides the current domain as the past domains to mine knowledge. Notice that for a current domain, only its domain data and the automatically accumulated knowledge will be used, and no other extra domain data will be available, which follows the lifelong learning experimental setting from existing works~\cite{chen2015lifelong,wang2016mining}. We empirically set the minimum support to 5 for frequent opinion word mining. We set the maximum iterations $m$ to 10 and the number of words $l$ to learn in each iteration to 50. A bigger $l$ makes the learning faster as it considers more words in one iteration, whereas a smaller $l$ makes the learning slower. We will show the effect of $m$ in Section~\ref{sec:analysis}.


\subsection{Quantitative Evaluation}
We use accuracy as the metric, as our task is a binary classification problem and the distribution of aspect-opinion words is nearly balanced (close to 3:2 in our labeled data). However, as it is hard to know the exact number of all related words (t-words) to a given target, we use the accuracy@$n$ (acc@$n$) as our evaluation measure, where $n$ is set to 50, 100, and 150. Given a target, we first obtain its t-words (nearest neighbors in the semantic space), and then manually label words as opinion or aspect (non-opinion) words. With the annotation obtained, we apply every trained candidate model to classify those top $n$ words to calculate its acc@$n$. 

\begin{figure*}[t]
\label{fig:acc50}
\centering
\subfloat[Acc@150 for all models and targets]{
\includegraphics[scale=.28]{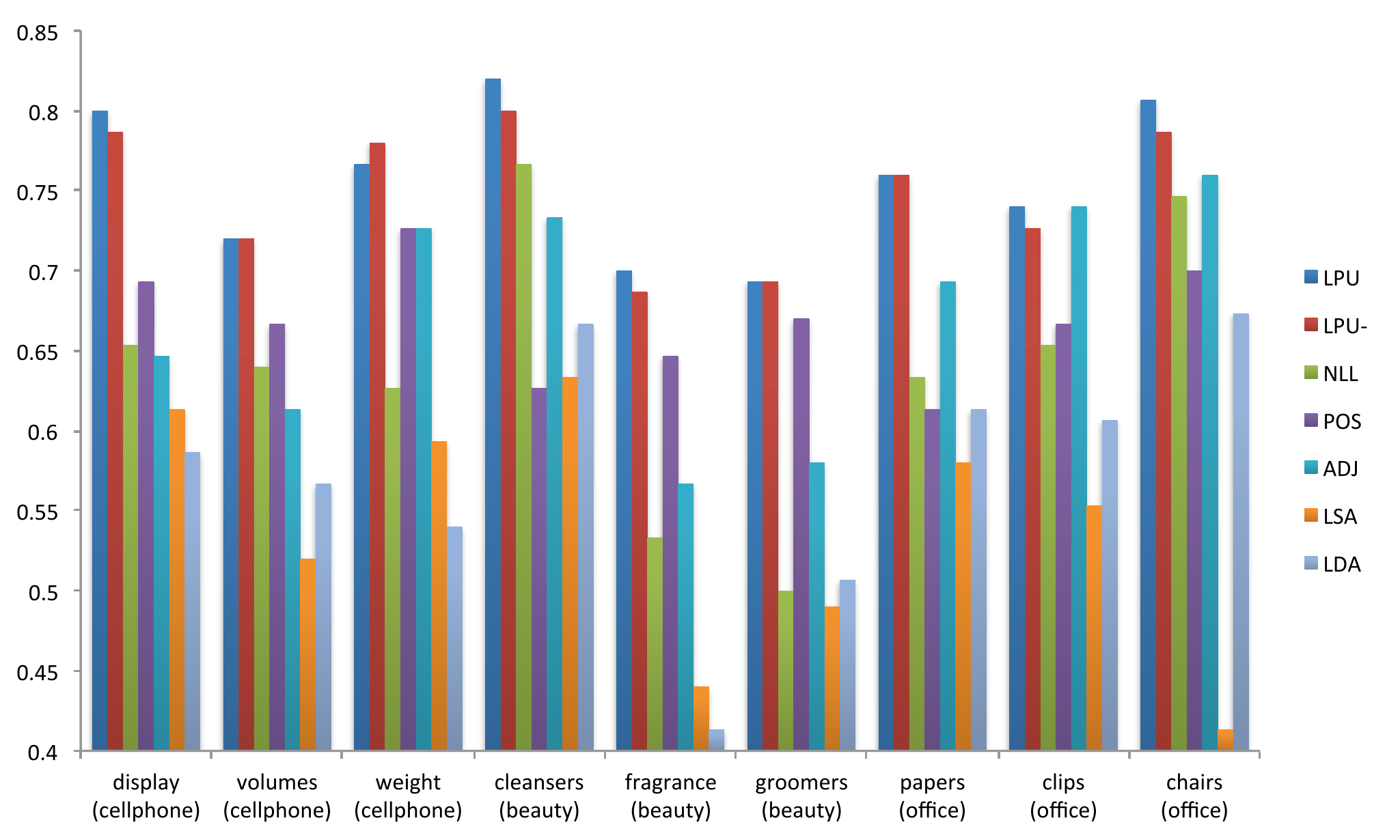}
\label{fig:acc150}
}
\subfloat[Acc@100 for all models and targets]{
\includegraphics[scale=.28]{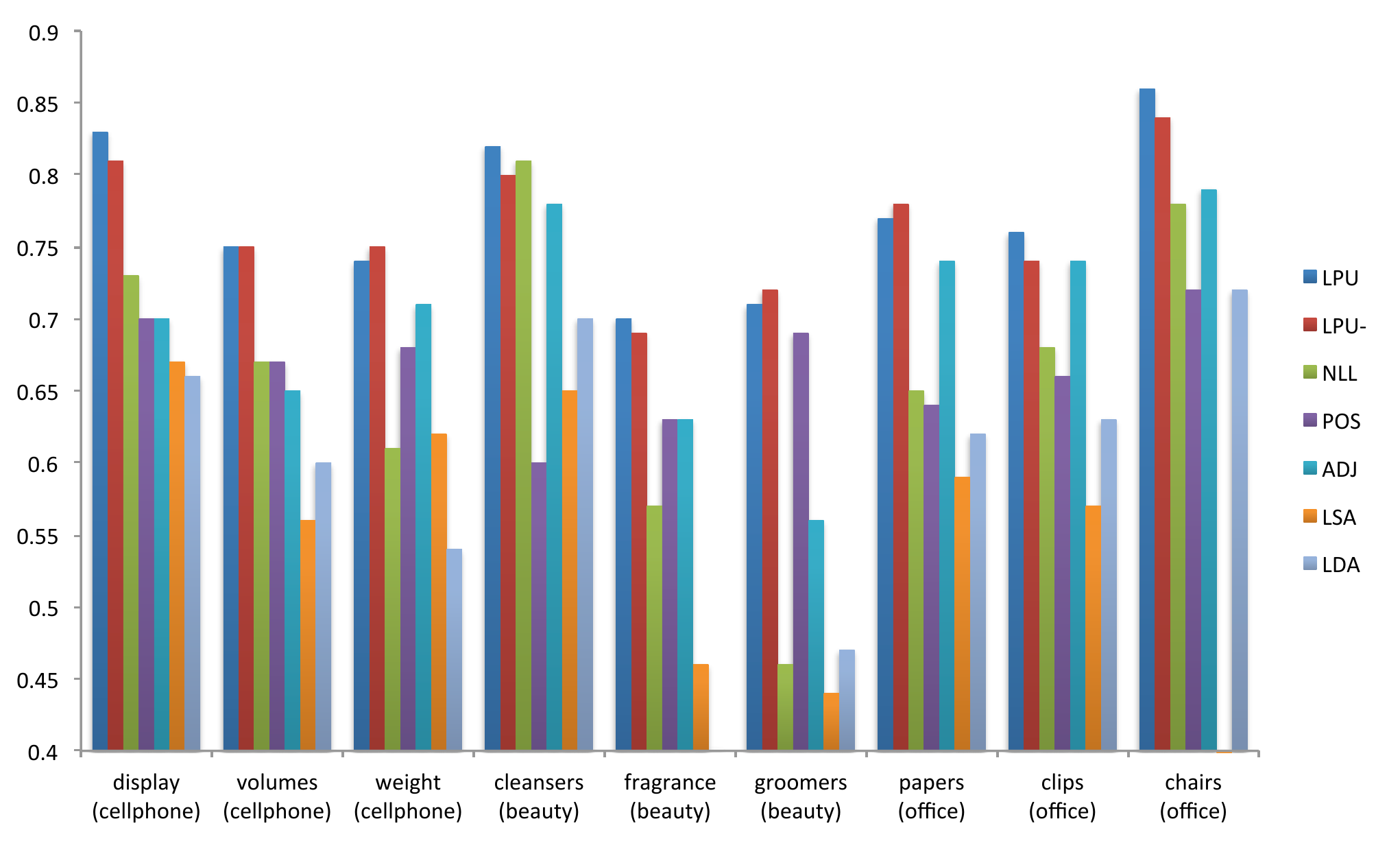}
\vspace{-10mm}
\label{fig:acc100}
}
\subfloat[Acc@50 for all models and targets]{
\includegraphics[scale=.28]{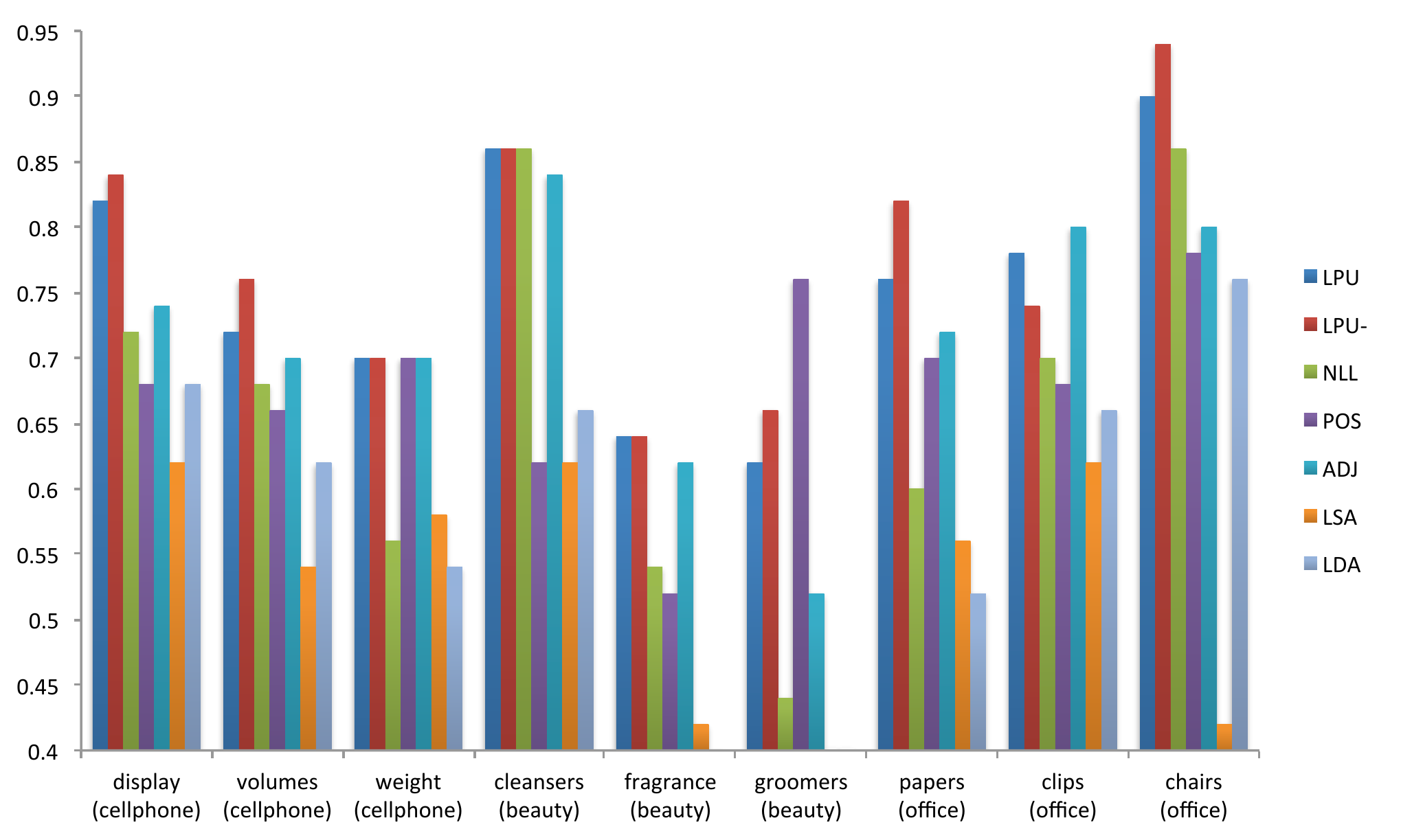}
\label{fig:acc50}
}
\vspace{-1.5mm}
\caption{Acc@n for all models and targets.}
\vspace{-3mm}
\end{figure*}

The results are reported in Figure~\ref{fig:acc150},~\ref{fig:acc100} and~\ref{fig:acc50}. Based on them, we have the following observations:

\begin{enumerate}[topsep=0pt,leftmargin=*]
\item LPU and LPU- outperform other baselines markedly. LPU improves the best baseline results by 8.29\%, 7.11\% and 4.00\% in acc@150, acc@100, and acc@50. Likewise, LPU- improves the best baseline results by 7.55\%, 6.44\% and 5.77\% in acc@150, acc@100, and acc@50. They demonstrate the effectiveness of lifelong learning. 

 \item LPU achieves better performance than LPU- in acc@150, acc@100 but is inferior to LPU- in acc@50. This indicates that LPU is more accurate by considering a big $n$ (more t-words), but LPU- could be more suitable if we only focus on the top-ranked words.

\item Among others baselines, we observe that NLL and POS perform the best. While POS explicitly reflects the contextual syntax, it is worth noting that the neural word embeddings used in NLL is also implicitly learned from word-context matrix~\cite{levy2015improving}. This implies that the syntactic information is very useful for word separation. 

\end{enumerate}

\renewcommand{\tabcolsep}{1.5pt} 
\begin{table}[t]
\tiny
\begin{center}
\begin{tabular}{|c|c|c|}
\hline
\multicolumn{3}{|c|}{\rule{0pt}{5pt} \textbf{Target: Volumes (Domain: CellPhone)}} \\
\hline
\textbf{Model} & \textbf{Aspect}	&	\textbf{Opinion} 	\\
\hline
LPU & volumes, bass, \wtw{undistorted}, volume, & muddiness, shrill, trebles, distortion, \\
&Gaga, pitches, sound, \wtw{harshness}, & hissy, loud, sibilance, thumping, \\
&cymbals, treble, eq, conf, & thump, highs, soundstage, midrange, \\
\hline
LPU- & bass, volume, Gaga, pitches, & \wtw{volumes}, muddiness, undistorted, shrill,\\
& sound, cymbals, treble, conf, & trebles, distortion, hissy, loud, \\
& Mids, reproduces, bitrate, Highs,& sibilance, harshness, thumping, eq, \\
\hline
NLL  &volumes, \wtw{muddiness}, bass, \wtw{undistorted}, & shrill, distortion, hissy, loud,\\
&  trebles, volume, Gaga, pitches, &  sibilance, midrange, \wtw{equalization}, tinny,\\
&  sound, \wtw{harshness}, cymbals, treble, &  distorted, piercing, muddy, louder, \\
\hline
POS & volumes, \wtw{muddiness}, trebles, Gaga, & \wtw{bass}, undistorted, shrill, distortion,   \\
&  pitches, \wtw{sibilance}, cymbals, Mids,  & hissy, \wtw{volume}, loud, \wtw{sound},  \\
& LiveAudio, highs, \wtw{reproducing}, Highs, &  harshness, \wtw{treble}, thumping, \wtw{eq}, \\
\hline
ADJ & volumes, \wtw{muddiness}, bass, trebles, & undistorted, shrill, distortion, hissy,  \\
&  volume, Gaga, pitches, sound,  & loud, \wtw{treble}, loudest, tinny,  \\
&  \wtw{sibilance}, \wtw{harshness}, cymbals, \wtw{thumping}, & distorted, \wtw{Treble}, resonant, muddy,\\
\hline
\end{tabular}
\end{center}
\vspace{-3mm} 
\caption{\label{tab:example1} Results for target \textit{volumes} in domain cellphone. Incorrect sentiment words are italicized and marked in red.}
\vspace{-2mm}
\end{table}

\subsection{Qualitative Evaluation}
This subsection shows some example results in Table~\ref{tab:example1}. Since LSI and LDA have much poorer performances than others, we do not include their results here. The represented words are the top predicted aspect words and top predicted opinion words from the t-words of a given target. Incorrect opinion words are italicized and marked in red.
As we can see, LPU and LPU- better distinguish aspect and opinion words. For example, the opinion word ``muddiness'' is extracted by them but not by other models. 
POS identifies many wrong opinion words like ``sound'' and ``volume''.  Although ADJ is good at extracting adjective opinion words, it misses other opinion words like ``muddiness'', ``sibilance'' and ``thumping''. NLL also misses many opinion words like ADJ.

\renewcommand{\tabcolsep}{1.5pt} 
\begin{table}[t]
\tiny
\begin{center}
\begin{tabular}{|c|c|c|}
\hline
\multicolumn{2}{|c|}{\rule{0pt}{5pt} \textbf{Topic: Skin (Domain: Beauty)}} \\
\hline
\textbf{Aspect}	&	\textbf{Newly-Identified Opinion} 	\\
\hline
face, skin, use, acne (-b), using,& dry (-c,d), rid (-c,d), oily (-c,d),   \\
 just (-b), wash, feel (-b), make,  &  drying (-a,c,d), mild (-c,d), notice (-c,d),  \\
day, product (-b), lotion & huge (-d), new (-c,d), ok (-c,d), younger (-c) \\
\hline
\end{tabular}
\end{center}
\vspace{-3mm} 
\caption{\label{tab:example3} Topic about skin in domain beauty. The ``-'' symbol indicates that the models following it do not identify the word. }
\vspace{-2.5mm}
\end{table}

\renewcommand{\tabcolsep}{1.5pt} 
\begin{table}[t]
\tiny
\begin{center}
\begin{tabular}{|c|c|c|}
\hline
\multicolumn{2}{|c|}{\rule{0pt}{5pt} \textbf{Topic: Headset (Domain: CellPhone)}} \\
\hline
\textbf{Aspect}	&	\textbf{Newly-Identified Opinion} 	\\
\hline
headset, sound, quality (-b),  & really (-c,d), long (-a), low (-d), \\
bluetooth, adapter, hear (-b), ear, &  away (-c,d), high (-d), short, quite (-c,d),\\
 volume, headsets, way (-b) &  ok, idea (-c,d), close (-c,d)  \\
\hline
\end{tabular}
\end{center}
\vspace{-3mm} 
\caption{\label{tab:example4} Topic about headset in domain cellphone. The ``-'' symbol indicates that the models following it do not identify the word.}
\vspace{-3mm}
\end{table}

\subsection{Further Analysis}
\label{sec:analysis}
In order to further evaluate the generality and extensibility of our proposed approach, we applied it to another popular aspect extraction technique, topic modeling. Specifically, we run LDA~\cite{blei2003latent} for topic generation and then use our algorithm to separate aspect words and opinion words. It also produces reasonably good results as shown in Table~\ref{tab:example3} and ~\ref{tab:example4}. Notations in ``-(a, b, c, d)'' will be explained below.

We also investigated the effect of alleviating FP error propagation in LPU. We denote three types of iterative-learning models as (a), (b), and (c), and they learned with 10 iterations by considering their newly-identified opinion words as positive examples: (a) LPU, using Alg.~1; (b) A PU model selecting its predicted positive examples ($prob>0.5$) in the current iteration as P for the next iteration; (c) A PU model that always combines the newly-predicted positive examples with the initial lexicon-based positive examples, without using the constraints in LPU. We also denote NLL as the model (d) for comparison purposes. It does not learn iteratively.

We now take a further look at Table~\ref{tab:example3} and ~\ref{tab:example4}. The ``-'' symbol indicates that the models following it do not identify the word. Note that here the opinion words from lexicon $P$ are excluded so we can see how those models perform in classifying the unlabeled words $U$. We observe: 1. Model (c) misses many interesting opinion words like model (d), which indicates that its positive examples remain very similar during all iterations, i.e., it does not learn many new positive examples; 2. Model (b) mis-classifies many aspect words as opinion words as its FP errors propagate iteratively, i.e., the model is confused by the newly-added false positive examples. 3. Model (a), which is LPU, works robustly well. 

\begin{figure}[t]
\begin{center}
\centerline{\includegraphics[scale=.40]{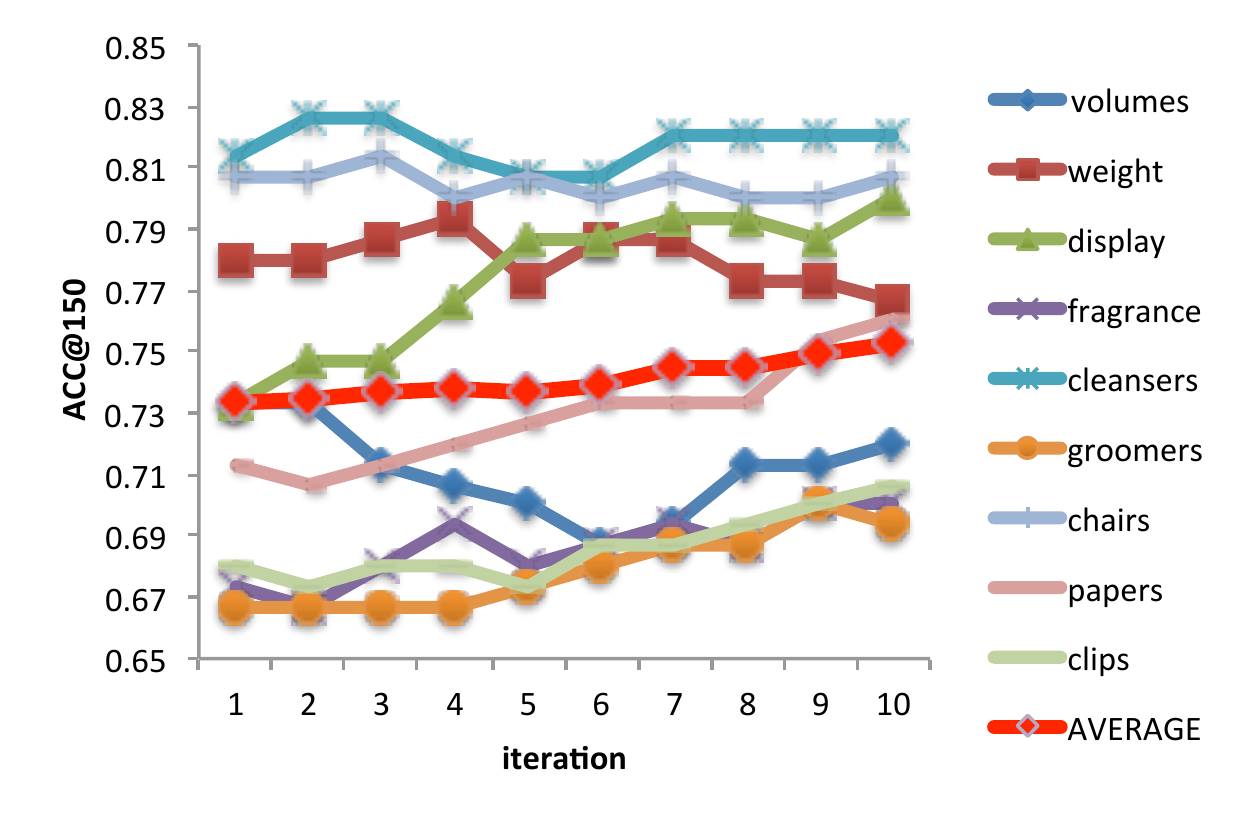}}
\vspace{-3mm}
\caption{\label{fig:iterative} Iterative learning of LPU}
\end{center}
\vspace{-9mm}
\end{figure}

We also report the effect of iterative learning of LPU with quantitative scores in Figure~\ref{fig:iterative}. The line in red color shows the averaged acc@150 scores in our labeled data. We can see the effectiveness and stability of LPU.

\section{Conclusion}
This paper discussed the problem of disentangling t-opinion words and t-aspect words from the grouped t-words for fine-grained target-based sentiment analysis (FTSA). We formulated this problem in a PU learning setting and incorporated the lifelong learning idea to overcome the drawback of error propagation in PU learning. To achieve this, a novel lifelong PU learning (LPU) model was proposed. Our experimental results using real-world data demonstrated its effectiveness.

\bibliographystyle{named}
\bibliography{main}

\end{document}